\documentclass[acmsmall]{acmart}

\AtBeginDocument{%
  }

\usepackage{natbib}
\usepackage{hyperref}  

\usepackage{amsmath} 

\hypersetup{
colorlinks   = true,
citecolor    = blue,
urlcolor    =  blue
}

\renewcommand\footnotetextcopyrightpermission[1]{} 
\begin{document}

\title{Unsupervised Bilingual Lexicon Induction for Low Resource Languages}

\author{Charitha Rathnayake}

\email{charitha.18@cse.mrt.ac.lk}
\orcid{0009-0008-3086-4623}

\author{P.R.S. Thilakarathna}
\authornote{Both authors contributed equally to this research.}
\email{ridmisameera.18@cse.mrt.ac.lk}
\orcid{0009-0003-5719-9540}

\author{Uthpala Nethmini}
\authornotemark[1]
\email{jasenpathiranage.18@cse.mrt.ac.lk}
\orcid{0009-0002-4252-7727}
\affiliation{%
  \institution{University of Moratuwa}
  \city{Katubedda}
  \country{Sri Lanka}
}

\author{Rishemjith Kaur}
\email{rishemjit.kaur@csio.res.in}
\orcid{}
\affiliation{%
  \institution{Central Scientific Instruments Organisation}
  \country{India}}

\author{Surangika Ranathunga}
\email{s.ranathunga@massey.ac.nz}
\orcid{0003-0701-0204}
\affiliation{%
  \institution{School of Mathematical and Computational Sciences, Massey University}
  \city{Auckland}
  \country{New Zealand}}

\renewcommand{\shortauthors}{Rathnayake et al.}

\begin{abstract}
Bilingual lexicons play a crucial role in various Natural Language Processing tasks. However, many low-resource languages (LRLs) do not have such lexicons, and due to the same reason, cannot benefit from the supervised Bilingual Lexicon Induction (BLI) techniques. To address this, unsupervised BLI (UBLI) techniques were introduced. A prominent technique in this line is structure-based UBLI. It is an iterative method, where a seed lexicon, which is initially learned from monolingual embeddings is iteratively improved. There have been numerous improvements to this core idea, however they have been experimented with independently of each other. In this paper, we investigate whether using these techniques simultaneously would lead to equal gains. We use the unsupervised version of VecMap, a commonly used structure-based UBLI framework, and carry out a comprehensive set of experiments using the LRL pairs, English-Sinhala, English-Tamil, and English-Punjabi. These experiments helped us to identify the best combination of the extensions. We also release bilingual dictionaries for English-Sinhala and English-Punjabi.
\end{abstract}




\received{20 February 2007}
\received[revised]{12 March 2009}
\received[accepted]{5 June 2009}

\maketitle

\section{Introduction}
A bilingual lexicon (aka bilingual dictionary) consists of a list of words in one language, along with the corresponding translations in another language. Bilingual lexicons have long been used to improve the performance of other Natural Language Processing (NLP) tasks such as Machine Translation~\cite{artetxe2018unsupervised}, Information Retrieval~\cite{vulic2015monolingual}, and cross-lingual Named Entity Recognition~\cite{mayhew2017cheap}. They are specifically useful in the context of Low-Resource Languages (LRLs)~\cite{miao2024enhancing, yong2024lexc, mohammed2023building, keersmaekers2023word, chaudhary2020dict, fernando2023exploiting, farhath2018integration}. 

However, due to the data scarcity of LRLs, many of them may not have manually developed bilingual lexicons. As a solution, researchers have proposed ways to automatically induce bilingual lexicons, which is known as Bilingual Lexicon Induction (BLI)~\cite{shi2021bilingual}. Early research used techniques such as those that exploit the similarity of word co-occurrences and the similarity of morphological structures of words~\cite{haghighi2008learning}. However, starting from the pioneering work of~\citet{mikolov2013exploiting}, techniques that make use of word embeddings have been at the forefront of BLI. 

Techniques such as the one presented by ~\citet{mikolov2013exploiting} are supervised, meaning that they require a bilingual lexicon (seed lexicon) to train the BLI model. However, many LRLs may not have such initial bilingual lexicons. As a solution, unsupervised BLI (UBLI) techniques have been introduced. These techniques rely solely on monolingual data. UBLI techniques can be broadly categorized into two: joint learning techniques and post-alignment techniques. In joint learning techniques, a neural network model is trained using monolingual data belonging to the two languages such that the resulting model captures the cross-lingual representations~\cite{cao2016distribution}. 

Post-alignment techniques, which have been more commonly used, start from the monolingual embeddings of the two languages and learn a mapping across their distributions. Post-alignment techniques can be further categorized as adversarial techniques and structure-based methods~\cite{ren2020graph}. Adversarial methods suffer from the limitations of the underlying adversarial models they use. In contrast, structure-based methods have shown to be more flexible and robust, and have been the state-of-the-art for UBLI~\cite{ren2020graph}. Most structure-based methods are iterative~\cite{aldarmaki2018unsupervised, hoshen2018non} - meaning that a seed lexicon is initially created, which is used to learn an optimal mapping between the embedding spaces of the two languages. A new lexicon is induced using the aligned embeddings, which is again used to further improve the alignment. This process is iteratively done until convergence. 

It is possible to improve structure-based UBLI by improved embedding generation techniques~\cite{ormazabal2021beyond, nishikawa2021data, cao2023dual, zhang2021combining}, embedding pre-processing techniques~\cite{vulic2020improving, artetxe2018uncovering, cao2023bilingual} and embedding initialization techniques~\cite{artetxe2018robust, li2020simple, ren2020graph, cao2021word, feng2022crosslingual}. However, these techniques have been experimented independently of each other, and it is not clear whether they would be still effective if simultaneously used. 

In order to answer this question, we select the commonly used structure-based UBLI framework introduced by \citet{artetxe2018robust} and extended it with a combination of the aforementioned techniques. This unsupervised framework was designed by extending their supervised self-learning framework VecMap~\cite{artetxe2017learning}. 
We carry out an extensive set of experiments for three LRL pairs, English-Sinhala (EnSi), English-Tamil (EnTa) and English-Punjabi (EnPa). 


The contributions of this paper are as follows:
\begin{itemize}   
    \item We carry out an extensive set of experiments on the combinations of different extensions to the structure-based UBLI systems using the unsupervised VecMap framework (UVecMap) and identify the most effective combination.
    \item We release new human-curated bilingual lexicons for English-Sinhala and English-Punjabi.
\end{itemize}

The rest of the paper is organized as follows. Section 2 discusses the UVecMap Framework, and its subsequent extensions, as well as BLI for LRLs. Section 3 discusses how the various extensions on UVecMap were implemented. Section 4 discusses the experiment setup and Section 5 discusses the results. Finally, Section 6 concludes the paper with a look into the future.

\section{Related Work}
\subsection{Structure-based UBLI with VecMap}
\label{sec:uvecmap}
\begin{figure}[h]
  \centering
  \includegraphics[width=0.4\linewidth]{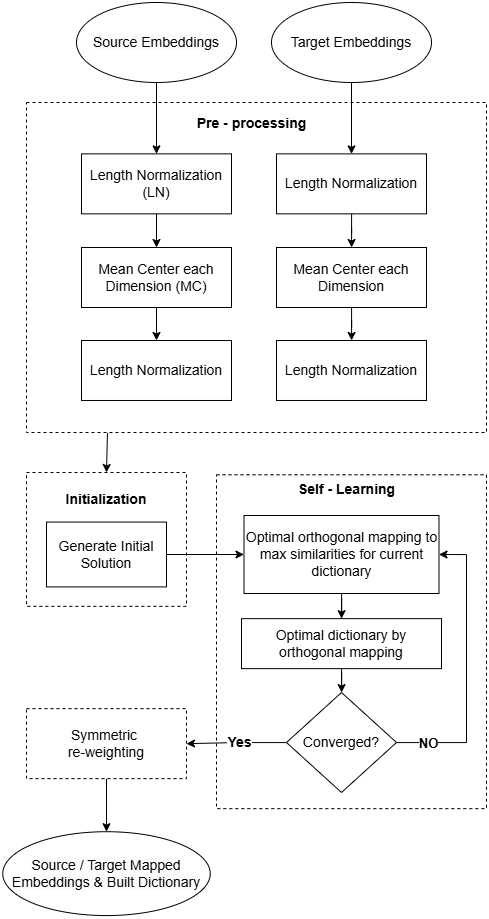}
  \caption{Schematic diagram of the UVecMap framework}
  \label{fig:vecmap}
  \Description{vecmap model}
\end{figure}

The UVecMap framework is shown in Figure ~\ref{fig:vecmap}. Note that UVecMap assumes that pre-trained word embeddings are already available for the considered languages. During the normalization step, the embeddings are length normalized, mean centered across each dimension, and again length normalized.  

In the second step, embeddings are initialized in an unsupervised manner. To do this, an alternative representation of the normalized source and target embeddings is derived as follows: given that the normalized embeddings for the source and target are $X$ and $Z$ respectively, similarity matrix $M_X$ corresponding to the source and the similarity matrix $M_Z$ corresponding to the target are first derived. Ideally, if the two embedding spaces were perfectly isometric\footnote{Two graphs that contain the same number of graph vertices connected in the same way are said to be isomorphic~\cite{sogaard2018limitations}. Isometry is isomorphism over matrix spaces.}, $M_X$ and $M_Z$ would be equivalent, apart from a permutation of their rows and columns. However, this is rarely the case in practice. To approximate alignment, each row in $M_X$ and $M_Z$  are independently sorted and their square root is calculated. This provides the initial dictionary.

During the self-learning step, the algorithm iteratively refines the cross-lingual embeddings until convergence by performing a sequence of optimization tasks. First, an optimal orthogonal mapping that maximizes the similarities for the current dictionary (D) is derived. For the $i^{th}$ vocabulary item $X^*_i$ in the source language and the $j^{th}$ vocabulary item $Z^*_j$ in the target language, the optimal solution is determined using the singular value decomposition (SVD) of $X^T$DZ, where $USV^T = X^T$DZ. The resulting mappings are given by $W_X = U$ and $W_Z = V$. Then, a new optimal dictionary is computed over the similarity matrix of the mapped embeddings, defined as $XW_X$$W^T_Z$$Z^T$. This dictionary is typically generated through nearest-neighbor retrieval from source to target embeddings ~\cite{artetxe2018robust}. Finally, both source and target embeddings are re-weighted. 

~\citet{artetxe2017learning} also introduced some additional improvements to this framework. First, dictionary induction was treated as a stochastic process, meaning that only a subset of the elements with probability $p$ are kept in the similarity matrices. Secondly, the dictionary size is limited to the top $k$ frequent words\footnote{~\citet{artetxe2018robust} used k = 20,000.}. Thirdly, to derive the dictionary from the aligned spaces, Cross-domain Similarity Local Scaling (CSLS)~\cite{lample2018word} is used, instead of the typically used nearest-neighbor retrieval. Finally, the dictionary is induced in both ways - from source to target and target to source. 

\subsection{Improvements to Structure-based BLI}
\label{sec:Improvements_to_UVecMap}
These improvements have been applied to one of the following components: word embedding creation, embedding pre-processing, and dictionary initialization. 

\paragraph{Word Embedding Creation}: ~\citet{ormazabal2021beyond} relaxed the assumption that pre-trained word embeddings are available for both source and target languages. They fixed the target embeddings and learned the aligned embeddings of the source from scratch. They used UVecMap to build the initial dictionary, which is used to constrain the source language embeddings to be aligned with the target language embeddings.~\citet{nishikawa2021data} showed that rather than deriving word embeddings from monolingual corpora, using a synthetic parallel corpus generated from unsupervised Machine Translation improves the performance of UVecMap. During the creation of word embeddings such as Word2Vec ~\cite{tomas2013efficient}, two types of embeddings are created- word embeddings and context embeddings. All the techniques discussed above used the word embedding part only. In contrast,~\citet{cao2023dual} used both word and context embeddings. The UVecMap model was originally designed to generate mappings using static word embedding models and does not inherently support embeddings with contextual representations. To address this limitation, ~\citet{zhang2021combining} proposed a method that combines static word embeddings with contextual representations to improve alignments and enhance Bilingual Lexicon Induction (BLI) results. Their approach employs FastText ~\cite{armand2016bagoftricks} embeddings for static representations and incorporates XLM ~\cite{lample2019crosslingual} and mBART ~\cite{liu2020multilingual} for contextual embeddings.

\paragraph{Embedding Pre-processing}: ~\citet{vulic2020improving} improved the quality of input monolingual spaces before using them in cross-lingual alignment. Their method incorporates a linear transformation approach proposed by ~\citet{artetxe2018uncovering}, which is controlled by a single parameter that adjusts the similarity order of the input embedding spaces. This linear transformation is applied to both monolingual spaces, enhancing the model's ability to capture different aspects of language.~\citet{cao2023bilingual} integrated the structural features from the source language embeddings into their corresponding features in the target language embeddings and vice versa. In other words, the value of a source side feature gets modified based on the corresponding one from the target side, and vice versa. They term this process `embedding fusion'. The objective of embedding fusion is to increase the isomorphism of the source and target spaces. ~\citet{cao2021word} introduced a transformation-based approach that applies rotation and scaling operations to monolingual embeddings, aiming to improve isomorphism between embedding spaces of different languages.
 
\paragraph{Initialization}: While it is always possible to use the generated word embeddings as they are, these embeddings often contain noise, which can negatively affect accuracy. To address this issue, ~\citet{li2020simple} performed dimensionality reduction of the original word embeddings in the initialization step. This method iteratively reduces the dimensionality of the embeddings using Principal Component Analysis (PCA), starting from the original dimension to the required target dimension. During each iteration, the algorithm generates a dictionary by computing the nearest neighbors of the k most frequently occurring words. The generated dictionaries are then compared to ensure that the final dictionary is the most accurate across all evaluated dimensionalities.~\citet{ren2020graph} introduced a graph-based solution for the initialization step. They constructed a graph using the monolingual embeddings of each language, where vertices represent words. Next, a subset of these vertices (called cliques) are extracted in such a manner that every two distinct vertices are adjacent. Then, embeddings of these cliques are calculated and the clique embeddings are mapped across the source and target graphs.  Central words of the aligned cliques are considered for the seed dictionary. 
~\citet{feng2022crosslingual} proposed a novel cross-lingual feature extraction (CFE) approach. They defined semantic features of words based on their relevance to contextual words, and quantified using character-level distances within monolingual corpora. Semantic vectors were constructed by selecting the most relevant contextual words, capturing language-independent textual features. The cross-lingual features were then combined with pre-trained embeddings.

\subsection{UBLI for Low-resource Languages}
~\citet{besacier2014automatic} defined an LRL as a ``language that lacks a unique writing system, lacks (or has) a limited presence on the World Wide Web, lacks linguistic expertise specific to that language, and/or lacks electronic resources such as corpora (monolingual and parallel), vocabulary lists'', and so on. Given that the success of NLP tools for a language depends on the language resource (raw, as well as annotated text data) availability, NLP researchers identify LRLs
considering the availability of text data and NLP tools as the criteria~\cite{joshi2020state, ranathunga2022some}. Accordingly, languages in the world have been categorized into 6 classes, with Class 0 being the least resourced\footnote{We use ~\citet{ranathunga2022some}'s language categorization. Language list can be found at \url{https://github.com/NisansaDdS/Some-Languages-are-More-Equal-than-Others/tree/main/Language_List/Language_Classes_According_To/DataSet_Availability} }. In this work, we consider languages belonging to classes 0-3 as LRLs. 

While BLI techniques can be applied to any language pair, not many have been tested on LRLs. We believe this is due to the lack of evaluation datasets.  The most commonly used dataset is the MUSE dataset~\cite{lample2018word}, which has lexicons for over hundreds of languages, including some LRLs. However, this dataset has been created automatically without human validation and have been reported to be of poor quality~\cite{yang2019maam}. PanLex~\cite{baldwin2010panlex} is another dataset that covers a large number of languages, however, this has also been automatically created.  Some other datasets created for LRLs have also been automatically created~\cite{glavas-etal-2019-properly, bafna2023simple, wickramasinghe2023sinhala}. For Indic languages, the IndoWordNet\footnote{\url{https://www.cfilt.iitb.ac.in/indowordnet/}} has been a good source to extract the evaluation dictionary.~\citet{pavlick2014language} present human created bilingual dictionaries for 100 languages, however, each language pair has only 10 entries. Some other research that created evaluation lexicons for LRLs have not publicly released them~\cite{liyanage2021bilingual}, thus impeding the progression in the field~\cite{ranathunga2024shoulders}.

\section{Methodology}
\begin{figure}[h]
  \centering
  \includegraphics[width=\linewidth]{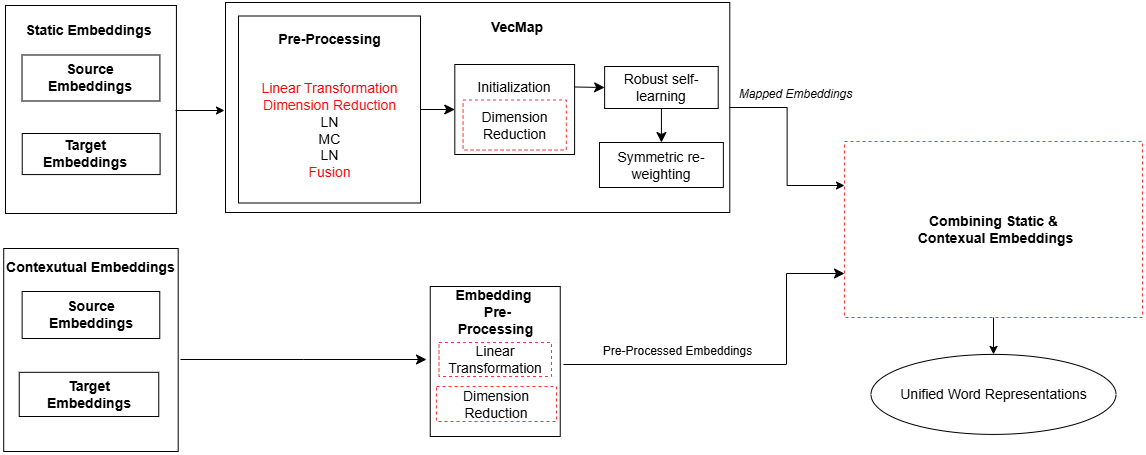}
  \caption{Improved UVecMap Framework. LN-Length Normalization, MC-Mean Center}
  \label{fig:impr_vecmap}
\end{figure}

As discussed in Section~\ref{sec:Improvements_to_UVecMap}, there have been many improvements to the core idea of structure-based UBLI. However, all those techniques have been individually experimented with. In order to verify their effectiveness when simultaneously used, we carried out a series of experiments. We use the UVecMap framework as the baseline structure-based UBLI system. 
Figure~\ref{fig:impr_vecmap} shows the UVecMap framework extended with some of the techniques discussed in Section~\ref{sec:Improvements_to_UVecMap}. Note that we considered only techniques for which the source code has been released. Components related to these improvements are shown in red color. Note that the step corresponding to dimensionality reduction during the pre-processing step is newly introduced by us. Below we discuss the technical details of these newly added components.

\subsection{Dimensionality Reduction}\label{sec:dim_reduction}
~\citet{li2020simple}'s technique discussed above integrates dimensionality reduction into the self-initialization step of the UVecMap framework. It is also possible to carry out this dimensionality reduction at the embedding pre-processing step. In our implementation, we experimented with both approaches independently of each other. We utilized Principal Component Analysis (PCA) to reduce the dimensionality of the embeddings. 


\subsection{Embedding pre-processing}\label{sec:pre_processing}
As discussed in Section ~\ref{sec:Improvements_to_UVecMap}, related work has discussed several ways of pre-processing word embeddings. Out of those, we experimented with linear transformation and embedding fusion techniques, on top of the pre-processing techniques already used in UVecMap. In addition to these, dimensionality reduction was also carried out as discussed earlier. 

\paragraph{Linear Transformation~\cite{vulic2020improving}: }Projection-based cross-lingual word embedding (CLWE) models typically learn a linear projection between two independently trained monolingual embedding spaces \(X\) and \(Z\) for source language \(L_s\) and target language \(L_t\), respectively. This alignment is guided by a word translation dictionary \(D\). This approach extends the traditional notion of similarity from first and second-order measures to higher 
\(n\)-th order similarities, enabling more robust alignment between the embedding spaces.

The first-order similarity matrix of the source language space \(X\) is calculated as \(M_1(X) = XX^T\). Similarly, the first-order similarity matrix for the target language space \(Z\) is determined. The second-order similarity can be expressed as \(M_2(X) = XX^TXX^T = M_1(M_1(X))\). Extending this, the \(n\)-th order similarity is defined as \(M_n(X) = (XX^T)^n\). 

~\citet{artetxe2018uncovering} proved that the \(n\)-th similarity transformation can be obtained using \(M_n(X) = M_1(XR_{\frac{n-1}{2}})\), where \(R_\alpha = Q\Delta^\alpha\), and \(\alpha\) is a hyperparameter. Here, \(Q\) and \(\Delta\) are obtained by the eigen-decomposition of \(X^TX = Q\Delta Q^T\), where \(Q\) is the orthogonal matrix with eigenvectors and \(\Delta\) is the diagonal matrix containing the eigenvalues. Separate hyperparameter values, \(\alpha_s\) and \(\alpha_t\), are defined for the source and target languages, respectively, to optimize the transformation for each language. The source embedding space is linearly transformed to \(X^l_{\alpha_s} = XR_{\alpha_s}\), where \(\alpha_s\) is the selected hyperparameter for the source language, and the target embedding space is similarly transformed as \(Z^l_{\alpha_t} = ZR_{\alpha_t}\), using \(\alpha_t\) for the target language. These transformations refine the original source embeddings, \(X\), and target embeddings, \(Z\), by incorporating the modified embeddings \(X^l_{\alpha_s}\) and \(Z^l_{\alpha_t}\), respectively.

\paragraph{Embedding Fusion ~\cite{cao2023bilingual}: }The fusion method proposed by ~\citet{cao2023bilingual} addresses the challenge of misaligned embedding spaces by improving the isomorphism between source embeddings (X) and target embeddings (Z) through rotation and joint scaling. It begins with the assumption of perfect isometry, where X and Z are related through a row permutation and an orthogonal rotation. Mathematically, this relationship is defined as $X=PZO$, where P is a permutation matrix containing 1s and 0s, and O is an orthogonal matrix. Using this assumption, the Singular Value Decomposition (SVD) of X and Z reveals the following relations:
\[
U_X = P U_Z, \quad S_X = S_Z, \quad V_X^T = V_Z^T O,
\]
where $U_XS_XV_X^T = X$ and $U_ZS_ZV_Z^T = Z$. This implies:
\[U_XS_X = PU_ZS_Z,\]
and consequently:
\[XV_X = PZV_Z\]
Here, $XV_X$ and $ZV_Z$ are the rotated embeddings, which are aligned across their dimensions up to a row permutation. Although the matrices P and O are unknown, this transformation places the row vectors of $XV_X$ and $ZV_Z$ in the same d-dimensional cross-lingual space.

Usually, embedding spaces are not perfectly isometric but are assumed to be approximately so. Although $XV_X$ and $ZV_Z$ may be roughly aligned, their singular values ($S_X$ and $S_Z$) often differ, especially for distant language pairs. Since $XV_X = U_XS_X$ and $ZV_Z = U_ZS_Z$, the embeddings are jointly scaled to align their singular value distributions and enhance their isomorphism. The scaling process is defined as $X'=XV_X\cdot\frac{\sqrt{S_Z}}{\sqrt{S_X}}, Z'=ZV_Z\cdot\frac{\sqrt{S_X}}{\sqrt{S_Z}}$,
where the operations are applied element-wise to the diagonal elements of $S_X$ and $S_Z$. The alignment ensures that $X^{\prime}$ and $Z^{\prime}$ are better suited for cross-lingual tasks, as their singular values are equalized and their geometric structures are aligned.

\subsection{Combining Static and Contextual Embeddings for Bilingual Lexicon Induction (CSCBLI) \label{sec:combining_method}}
~\citet{zhang2021combining} considered only FastText alongside XLM and mBART embeddings in their methodology. In our work, we extended their approach by incorporating both Word2Vec and FastText as static embeddings, alongside XLM-R ~\cite{alexis2020unsupervised} as the contextual embeddings. 
The proposed model consists of two primary steps: Unified Word Representation Space and Similarity Interpolation.

In the first step, the model constructs a unified word representation space that combines static word embeddings and contextual representations. Since the embedding spaces of the two languages are not perfectly isometric, some translation pairs remain distant even after mapping. To address this, a spring network adjusts the mapped embeddings, pulling translation pairs closer together to ensure that they become nearest neighbors.

The unified word representations, \( U_x \) and \( U_y \), are mathematically defined as:

\[
U_x = E'_x + \gamma_1 \odot F_x(A_x), \quad U_y = E'_y + \gamma_2 \odot F_y(A_y),
\]
where \( E'_x \) and \( E'_y \) are the mapped word embeddings, \(F_x(A_x)\) and \(F_y(A_y)\) are the weighted offsets produced by the spring networks \( F_x \) and \(F_y\) based on the contextual representation matrices \( A_x \) and \(A_y\), and \( \gamma_1 \) and \( \gamma_2 \) are weight vectors scaling the offsets. These weights are initialized as zero vectors, and during the training process, these parameters are updated by backpropagation to optimize the performance of the model.

The spring network comprises two layers. The first layer transforms the dimensionality of the contextual representation (\( d_0 \)) to match that of the static word embedding (\( d \)), as shown in the following equation:

\[
A_x^1 = \varphi(\theta_x^0(A_x)), \quad A_y^1 = \varphi(\theta_y^0(A_y)),
\]
where \( \varphi \) denotes the Tanh activation, and \( \theta \) represents the feedforward layers. The second layer refines the representation by mapping the output of the first layer into the final offset values.

\[
A_x^2 = \varphi(\theta_x^1(A_x^1)), \quad A_y^2 = \varphi(\theta_y^1(A_y^1)),
\]

The outputs \( A_x^2 \) and \( A_y^2 \) serve as offsets to correct deviations in the mapped word embedding space.

To optimize the spring network, contrastive training is used. Initially, the bilingual dictionary is generated using the bilingual word mappings based on static word embeddings. The model iteratively refines this dictionary by finding new translations for source words and updating the dictionary after each iteration. This process continues until the dictionary stabilizes, ensuring improved alignment between the source and target embeddings.

During inference, the model performs similarity interpolation between the unified representation space and the contextual space to compute final translation similarities. Both the unified word representation space and the mapped contextual representation space, as well as the lambda value are given as the inputs for the inference. Given a source word \(x\), the similarities are interpolated as follows:

\[
S = \cos(U_x, U_y) + \lambda \cos(A_x^0, A_y^0)
\]

where \(\lambda\) is the weight, and \(A_x^0\) and \(A_y^0\) are the mapped contextual representations.

\section{Experiment Setup}

\subsection{\textbf{Monolingual Data}}
\label{sec:monolingual_data}
Unsupervised BLI techniques require monolingual corpora to generate initial language-specific embeddings. While there are web-crawled corpora, those corresponding to low-resource languages are noisy~\cite{ranathunga2024quality}. Therefore, for Sinhala-English and Tamil-English pairs, we used the monolingual versions of the SiTa parallel corpus ~\cite{fernando2020data}. This corpus has been meticulously cleaned by humans. Similarly, the Bharath Parallel Corpus Collection (BPCC) ~\cite{gala2023indictrans2} was utilized for English-Punjabi. These corpora were tokenized to derive word lists.  

{Previous research suggested that keeping the full word list is not effective, due to the existence of rare words \cite{artetxe2018robust, feng2022crosslingual}. Therefore, the resulting word lists were filtered based on predefined minimum frequency $Min\_Freq$ threshold values.} The thresholds were carefully selected via an ablation study (see Section ~\ref{sec:experiment}) to balance coverage (the range of vocabulary captured) and accuracy (the quality of the embeddings). Increasing the frequency threshold typically reduces the number of words (lower coverage) but improves the quality of embeddings by focusing on more representative terms (higher accuracy). Conversely, lowering the threshold increases coverage by including more words, but it may dilute the quality of embeddings due to the inclusion of infrequent and less meaningful terms. 

\subsection{\textbf{Evaluation Dictionary}}
\label{sec:dic}
Even though the BLI technique is unsupervised, a human-curated dataset is needed to evaluate the performance of the models. Depending on resource availability, we prepared the evaluation dictionaries as follows.

\subsubsection{Sinhala-English}
\label{sec:EnSi_eval}

\textbf{Using a Machine Translation Tool: } We followed the method outlined by~\citet{zhang2017adversarial} to create the evaluation dictionary. Initially, word sets for English and Sinhala were extracted from the SiTa corpus as described in section ~\ref{sec:monolingual_data}. The English word list was then sorted by frequency, starting from the most frequent word and proceeding to the least frequent. Each selected English word was translated to Sinhala using a Machine Translation tool (we used Google Translate\footnote{https://translate.google.com/m}). The resulting Sinhala word was again back-translated to English using the same tool. From this output, the following word pairs were discarded: 
\begin{itemize}
    \item \textbf{Discrepancies}: Words with any inconsistencies between the original English word and its back-translated counterpart.
    \item \textbf{Out-of-Vocabulary}: Words with translations that did not match the target language vocabulary derived from the target language corpus.
    \item \textbf{Multi-Word Translations}: Words whose translations resulted in multi-word phrases.
    \item \textbf{Proper Nouns}
\end{itemize}

 All the challenges we encountered in creating this dictionary and the corresponding steps taken to address them are detailed in Table ~\ref{tab:issues_table}. 
 This process continued until we were able to compile a minimum of 1,500 word pairs for the evaluation set, aligning with the typical size of evaluation sets used in most previous research~\cite{zhang2021combining,feng2022crosslingual}. The final dictionary consists of 1,562 word pairs.

\begin{table}[h]
\centering
\caption{Issue in creating the bilingual lexicon using the translation tool \& Solutions}
\begin{tabular}{|c|}
\hline
 \includegraphics[width=\linewidth]{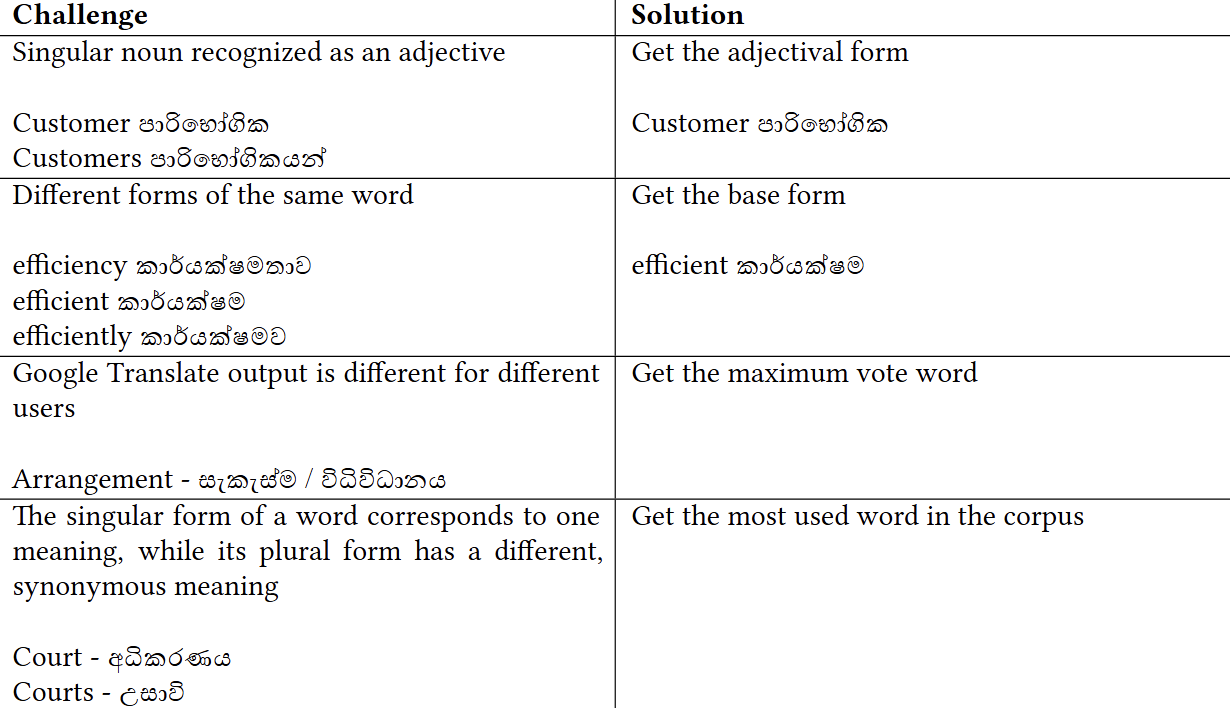} \\ \hline
\end{tabular}
\label{tab:issues_table}
\end{table}

\subsubsection{Tamil-English} 

\textbf{Using an existing Lexicon:} As described in the section ~\ref{sec:monolingual_data}, we constructed separate monolingual word sets for English and Tamil. Using an existing English-Tamil dictionary\footnote{https://education.nsw.gov.au/content/dam/main-education/teaching-and-learning/curriculum/multicultural-education/eald/eald-bilingual-dictionary-tamil.pdf}, we checked each word pair to verify whether both the English word and the corresponding Tamil word were present in our respective word lists. If both words were found in the lists, the pair was included in our evaluation dictionary. The final evaluation dataset comprises 1,155 word pairs extracted from the complete dictionary.

\subsubsection{Punjabi-English}
\textbf{Using IndoWordNet:}
 We used the existing English-Punjabi word pair lists available from IndoWordNet\footnote{ https://github.com/cfiltnlp/IWN-WordLists/tree/main/bilingual/English-Punjabi}. Initially, we removed multi-phrase Punjabi words from the lists. We further removed the English-Punjabi word pairs where either English or Punjabi word was absent from  BPCC Samanantar parallel corpora  ~\cite{gala2023indictrans2}. 
Next, we calculated the frequency of occurrence for each word in both English and Punjabi lists in the BPCC dataset. Words with a frequency of occurrence less than nine were excluded. Additionally, we removed words with multiple Punjabi translations to ensure a one-to-one correspondence between English and Punjabi words. During manual review, we identified certain translations that were not the most commonly used terms. These translations were also removed, along with the proper nouns. Following these pre-processing steps, the final dataset consisted of 1,391 English-Punjabi word pairs.

\subsection{Computing Environment}
Initial experiments using Word2Vec and FastText embeddings, each with 300 dimensions, were conducted on Google Colab. Subsequently, experiments with XLM-R embeddings, which have 1024 dimensions, were conducted in a GPU (Single NVIDIA Quadro M6000).

\subsection{Experiment Setup} 
\label{sec:experiment}
\textbf{Embedding Creation:} We carried out an ablation study to select threshold values for minimum frequencies, which resulted in the minimum frequencies being 8 for EnSi, and 6 for both EnTa and EnPa. The detailed results of the ablation study can be found in Tables ~\ref{tab:min_freq_enta}-\ref{tab:min_freq_ensi} of Appendix ~\ref{sec:appendix_threshold}. Based on these thresholds, Word2Vec and FastText static word embeddings, along with XLM-R contextualized embeddings, were generated for the selected words. Counts\footnote{Although the same word sets were provided, Word2Vec, FastText, and XLM-R generated embeddings only for words present in their respective vocabularies. Since Word2Vec and FastText were trained on the same corpus with different pre-processing steps, their SRC embedding counts are similar, while XLM-R differs due to its unique tokenization and coverage.} are shown in Table ~\ref{tab:embeddings}. 

\begin{table}[h]
    \centering
    \caption{Embedding Sizes for Language Pairs. SRC-Source Language, TRG-Target Language}
    \label{tab:embeddings}
    \begin{tabular}{|c|c|c|c|c|c|c|}
        \hline
        \textbf{language pair} & \multicolumn{2}{|c|}{\textbf{W2V}} & \multicolumn{2}{|c|}{\textbf{FASTTEXT}} & \multicolumn{2}{|c|}{\textbf{XLMR}} \\ \hline
        & \textbf{SRC} & \textbf{TRG} & \textbf{SRC} & \textbf{TRG} & \textbf{SRC} & \textbf{TRG} \\ \hline
        \textbf{EnSi} & 5101 & 6303 & 5101 & 6303 & 4845 & 6117  \\ \hline
        \textbf{EnTa} & 6034 & 10587 & 6034 & 10587 & 5758 & 10413 \\ \hline
        \textbf{EnPa} & 47783 & 68300 & 47783 & 68300 & 44296 & 65702 \\ \hline
    \end{tabular}
    
\end{table}

In addition, we needed to determine some parameters specific to the individual techniques we used. We integrated these techniques one at a time with UVecMap and then combined them in an iterative manner to determine the relevant hyper-parameters.

 \subsubsection{\textbf{Linear Transformation Method:}} \label{sec:linear_experiment} As discussed in Section ~\ref{sec:pre_processing}, tuning the hyperparameters \(\alpha_s\) and \(\alpha_t\) for the source and target languages, respectively, is crucial. However, due to resource constraints, rather than exploring the full parameter space, we manually defined a subset of the hyperparameter space based on the experiments conducted by~\citet{vulic2020improving}. Specifically, we applied post-processing to the corpora for both the source and target languages using hyperparameter values \([-0.5, -0.25, -0.15, 0, 0.15, 0.25, 0.5]\). Subsequently, we ran the UVecMap model for each pair of \(\alpha_s\) and \(\alpha_t\) values to generate cross-lingual word embeddings. These embeddings were evaluated against the evaluation dictionaries we created. This experiment was performed using both FastText and Word2Vec embeddings.  For the XLM-R embeddings, separate \(\alpha\) values were required when combined with Word2Vec and FastText. To generate the cross-lingual word embeddings for XLM-R, we used the CSCBLI model and evaluated these embeddings against the evaluation sets we developed. The selected \(\alpha\)-values for the source and target embeddings, identified after hyperparameter training, are summarized in Table ~\ref{tab:alpha_values}. Tables ~\ref{tab:alpha_ensi_fasttext} -~\ref{tab:alpha_enpj_w2v}, which present the hyperparameter training values for alpha selection for FastText and Word2Vec, are included in Appendix~\ref{sec:appendix_linear}.
 
 \begin{table}[h]
    \centering
    \caption{Alpha Parameter Values for EnSi, EnPa, and EnTa Embeddings. SRC- source language, TRG - target language.}
    \label{tab:alpha_values}
    \begin{tabular}{|c|c|c|c|c|c|c|c|c|}
        \hline
        & \multicolumn{2}{|c|}{\textbf{W2V}} & \multicolumn{2}{|c|}{\textbf{FASTTEXT}} & \multicolumn{2}{|c|}{\textbf{XLMR+W2V}} & \multicolumn{2}{|c|}{\textbf{XLMR+FASTTEXT}} \\ \hline
        & \textbf{SRC} & \textbf{TRG} & \textbf{SRC} & \textbf{TRG} & \textbf{SRC} & \textbf{TRG} & \textbf{SRC} & \textbf{TRG} \\ \hline
        \textbf{EnSi} & 0.15 & 0.25 & 0 & 0.25 & 0 & -0.5 & -0.15 & 0.25 \\ \hline
        \textbf{EnTa} & 0.15 & 0 & 0.15 & 0.15 & 0.15 & 0 & 0.15 & 0.15 \\ \hline
        \textbf{EnPa} & -0.25 & 0 & -0.15 & 0.25 & 0.25 & 0 & 0.25 & 0 \\ \hline
    \end{tabular}
    
\end{table}

  \subsubsection{\textbf{Dimension Reduction:}} \label{sec:effective_experimen}
As outlined in Section~\ref{sec:dim_reduction}, dimensionality reduction was applied in both the preprocessing and self-initialization stages of our implementation. Here, we specifically discuss its application during the preprocessing phase. The original dimensionalities of the embeddings are 300 for Word2Vec and FastText, and 1024 for XLM-R. To enhance computational efficiency and minimize redundancy, we applied dimensionality reduction using PCA, halving the original dimensions. This resulted in embeddings with 150 dimensions for Word2Vec and FastText, and 512 dimensions for XLM-R. These reduced embeddings formed a more compact and efficient representation while retaining key semantic features for subsequent tasks.

\section{Results and Discussion}
\label{sec:results}
The reported results for the conducted experiments followed the standard practice of using precision@k (Pr@k), with k = 1, as the evaluation metric. Pr@k represents the number of accurate translations found in the top k retrieved result set. All the experimented method combinations are listed in Table ~\ref{tab:experiments}, along with a code for easy reference thereafter. Experiment results are in Table ~\ref{tab:results}. 

\begin{table}[H]
    \caption{Overview of Experiment Combinations and Codes}
    \label{tab:experiments}
    \begin{tabular}{|p{10cm}|p{4cm}|}
        \hline
        \textbf{Experiment} & \textbf{Code} \\ \hline
        UVecMap (baseline) & M1 \\ \hline
        Effective Dim. Reduction + UVecMap & M2 \\ \hline
        Linear Transformation + UVecMap & M3 \\ \hline
        Linear Transformation + Effective Dim. Reduction + UVecMap & M4 \\ \hline
        Effective Dim. Reduction + Linear Transformation + UVecMap & M5 \\ \hline
        Iterative Dim. reduction + UVecMap & M6 \\ \hline
        Linear Transformation + Iterative Dim. Reduction + UVecMap & M7 \\ \hline
        UVecMap + Fusion & M8 \\ \hline
        Effective Dim. Reduction + UVecMap + Fusion & M9 \\ \hline
        Linear Transformation + UVecMap + Fusion & M10 \\ \hline
        Linear Transformation + Effective Dim. Reduction + UVecMap + Fusion & M11 \\ \hline
        Effective Dim. Reduction + Linear Transformation + UVecMap + Fusion & M12 \\ \hline
        Iterative Dim. reduction + UVecMap + Fusion & M13 \\ \hline
        Linear Transformation + Iterative Dim. Reduction + UVecMap + Fusion & M14 \\ \hline
        CSCBLI + UVecMap & M15 \\ \hline
        CSCBLI + Effective Dim.Reduction + UVecMap & M16 \\ \hline
        CSCBLI + Linear Transformation + UVecMap & M17 \\ \hline
        CSCBLI + Linear Transformation + Effective Dim. Reduction + UVecMap & M18 \\ \hline
        CSCBLI + Iterative Dim. reduction + UVecMap & M19 \\ \hline
        CSCBLI + Linear Transformation + Iterative Dim. Reduction + UVecMap & M20 \\ \hline
    \end{tabular}
\end{table}

\begin{table}[h!]
    \centering
    \caption{pr@1 Results of Experiment Combinations. For each column, the best result is in boldface, second best is in italics. Gain over the baseline (M1) is given within brackets.}
    \label{tab:results}
    \renewcommand{\arraystretch}{1.2} 
    \setlength{\tabcolsep}{3pt} 
    \begin{tabular}{|p{3cm}|c|c|c|c|c|c|}
        \hline
        \textbf{Experiment} & \multicolumn{2}{|c|}{\textbf{EnSi}} & \multicolumn{2}{|c|}{\textbf{EnTa}} & \multicolumn{2}{|c|}{\textbf{EnPa}} \\ \hline
        & \textbf{W2V} & \textbf{FASTTEXT} & \textbf{W2V} & \textbf{FASTTEXT} & \textbf{W2V} & \textbf{FASTTEXT} \\ \hline
        \textbf{M1} & 31.49 & 27.48 & \textit{16.74} & 11.69 & 15.13 & 15.05 \\ \hline
        \textbf{M2} & 31.22$_{(-0.27)}$ & 27.74$_{(+0.26)}$ & 16.07$_{(-0.67)}$ & 0.11$_{(-11.58)}$ & 13.48$_{(-1.65)}$ & 14.81$_{(-0.24)}$ \\ \hline
        \textbf{M3} & \textbf{33.18}$_{(+1.69)}$ & 28.81$_{(+1.33)}$ & 16.4$_{(-0.34)}$ & 14.04$_{(+2.35)}$ & 15.36$_{(+0.23)}$ & \textit{16.22}$_{(+1.17)}$ \\ \hline
        \textbf{M4} & 32.11$_{(+0.62)}$ & 27.74$_{(+0.26)}$ & 14.49$_{(-2.25)}$ & 0$_{(-11.69)}$ & 14.03$_{(-1.1)}$ & 13.87$_{(-1.18)}$ \\ \hline
        \textbf{M5} & 31.58$_{(+0.09)}$ & 27.21$_{(-0.27)}$ & 15.84$_{(-0.9)}$ & 13.48$_{(+1.79)}$ & 14.26$_{(-0.87)}$ & 14.5$_{(-0.55)}$ \\ \hline
        \textbf{M6} & 32.83$_{(+1.34)}$ & 26.67$_{(-0.81)}$ & 16.52$_{(-0.22)}$ & 12.58$_{(+0.89)}$ & 15.13$_{(0)}$ & 15.91$_{(+0.86)}$ \\ \hline
        \textbf{M7} & 32.11$_{(+0.62)}$ & 27.3$_{(-0.18)}$ & 8.99$_{(-7.75)}$ & \textit{14.16}$_{(+2.47)}$ & 15.20$_{(+0.07)}$ & 15.36$_{(+0.31)}$ \\ \hline
        \textbf{M8} & 11.24$_{(-20.25)}$ & 23.19$_{(-4.29)}$ & 2.81$_{(-13.93)}$ & 0.11$_{(-11.58)}$ & 14.97$_{(-0.16)}$ & 15.28$_{(+0.23)}$ \\ \hline
        \textbf{M9} & 20.25$_{(-11.24)}$ & 26.32$_{(-1.16)}$ & 14.72$_{(-2.02)}$ & 13.37$_{(+1.68)}$ & 12.77$_{(-2.36)}$ & 13.71$_{(-1.34)}$ \\ \hline
        \textbf{M10} & 31.31$_{(-0.18)}$ & 27.03$_{(-0.45)}$ & 4.49$_{(-12.25)}$ & 0$_{(-11.69)}$ & 14.81$_{(-0.32)}$ & 15.36$_{(+0.31)}$ \\ \hline
        \textbf{M11} & 30.6$_{(-0.89)}$ & 19.27$_{(-8.21)}$ & 1.46$_{(-15.28)}$ & 0$_{(-11.69)}$ & 14.18$_{(-0.95)}$ & 14.26$_{(-0.79)}$ \\ \hline
        \textbf{M12} & 31.85$_{(+0.36)}$ & 26.58$_{(-0.9)}$ & 12.47$_{(-4.27)}$ & 0.11$_{(-11.58)}$ & 14.26$_{(-0.87)}$ & 13.32$_{(-1.73)}$ \\ \hline
        \textbf{M13} & 12.67$_{(-18.82)}$ & 23.28$_{(-4.2)}$ & 2.92$_{(-13.82)}$ & 0.67$_{(-11.02)}$ & 15.13$_{(0)}$ & 15.2$_{(+0.15)}$ \\ \hline
        \textbf{M14} & 32.29$_{(+0.8)}$ & 27.83$_{(+0.35)}$ & 5.17$_{(-11.57)}$ & 0$_{(-11.69)}$ & 14.97$_{(-0.16)}$ & 15.67$_{(+0.62)}$ \\ \hline
        \textbf{M15} & 31.07$_{(-0.42)}$ & 28.37$_{(+0.89)}$ & 16.13$_{(-0.61)}$ & 13.36$_{(+1.67)}$ & \textit{15.50}$_{(+0.37)}$ & 15.24$_{(+0.19)}$ \\ \hline
        \textbf{M16} & 30.42$_{(-1.07)}$ & 28.47$_{(+0.99)}$ & 16.73$_{(-0.01)}$ & 0.12$_{(-11.57)}$ & 14.29$_{(-0.84)}$ & 15.15$_{(+0.1)}$ \\ \hline
        \textbf{M17} & \textit{32.84}$_{(+1.35)}$ & \textbf{29.49}$_{(+2.01)}$ & \textbf{16.85}$_{(+0.11)}$ & \textbf{15.28}$_{(+3.59)}$ & \textbf{15.58}$_{(+0.45)}$ & \textbf{16.36}$_{(+1.31)}$ \\ \hline
        \textbf{M18} & 32.09$_{(+0.6)}$ & \textit{28.93}$_{(+1.45)}$ & 15.64$_{(-1.1)}$ & 0$_{(-11.69)}$ & 14.55$_{(-0.58)}$ & 14.98$_{(-0.07)}$ \\ \hline
        \textbf{M19} & 31.91$_{(+0.42)}$ & 28.37$_{(+0.89)}$ & 16.25$_{(-0.49)}$ & 12.76$_{(+1.07)}$ & \textbf{15.58}$_{(+0.45)}$ & 15.76$_{(+0.71)}$ \\ \hline
        \textbf{M20} & 32.19$_{(+0.7)}$ & 28.65$_{(+1.17)}$ & 9.38$_{(-7.36)}$ & \textbf{15.28}$_{(+3.59)}$ & 15.41$_{(+0.28)}$ & 15.67$_{(+0.62)}$ \\ \hline
    \end{tabular}
\end{table}

As for the baseline, performance of the two types of embeddings is not consistent: For EnSi, Word2Vec dominates over FastText across all the experiments. The same holds for Tamil. 
However, Word2Vec and FastText results remain on par for EnPa.

It appears that CSCBLI with Linear Transformation and UVecMap (M17) yielded the best results for both Word2Vec and Fasttext. The combinations of CSCBLI with Iterative Dimension Reduction and UVecMap (M19), as well as CSCBLI with Linear Transformation, Iterative Dimension Reduction and UVecMap (M20), also produced similar results to that of M17 for certain language embeddings. These combinations demonstrate the effectiveness of integrating multiple techniques to improve the quality of embeddings and alignment. The only exception is the result for EnSi with Word2Vec embeddings, where the best result was achieved by combining linear transformation with UVecMap (M3).

Note that in the EnTa FastText embeddings, some combinations result in the performance dropping to zero. This variation occurs because the mapping relies heavily on a few key embeddings. When these get modified by different techniques, the mapping can fail, leading to poor results.

Although we experimented with the fusion technique, the obtained accuracies, as detailed in Table ~\ref{tab:results} (M8-M14), were not satisfactory and fell short of the desired performance metrics.

\section{Conclusion}
The majority of research on BLI has predominantly focused on high-resource languages, leaving low-resource languages under-explored. Challenges such as the scarcity of parallel data, the nature of available datasets, and limitations in BLI and embedding techniques have resulted in suboptimal accuracy in previous studies. To address these issues, our research focused on enhancing the UVecMap model, a prominent unsupervised BLI framework. 
To achieve this, we employed a combination of techniques proposed by past research, in order to improve embedding creation, embedding pre-processing, and dictionary initialization. 
Our results show that combining multiple techniques over UVecMap provides the optimal BLI performance across three low-resource language pairs. The new bilingual lexicons we created for EnSi and EnPa can be found at our \href{https://github.com/CharithaRathnayake/BLI}{Github Repository}.

However, despite these improvements, our research has several limitations. One such limitation is the constraint of GPU memory when running the CSCBLI method. This limitation restricted the maximum count of embeddings that could be processed, which in turn affected the scope of the language pairs and the volume of data used. The approach we used also requires manually setting hyper-parameters, which can limit its scalability and applicability to other language pairs with different linguistic characteristics. Future work will focus on optimizing memory usage, automating the hyper-parameter tuning process, and applying our approach to a broader range of LRLs.
\begin{acks}
    We thank Anushika Liyanage for sharing the findings and insights from her prior work with us. We acknowledge the contribution of Gautam for the information provided in preparing the Punjabi lexicon.  
\end{acks}
\bibliographystyle{ACM-Reference-Format}
\bibliography{sample-base}

\appendix

\section{Linear Transformation Method}
\label{sec:appendix_linear}
Tables~\ref{tab:alpha_ensi_fasttext} - ~\ref{tab:alpha_enpj_w2v} present the results obtained during the hyperparameter tuning process, as described in Section ~\ref{sec:linear_experiment} .

\begin{table}[h]
    \centering
    \caption{Alpha Hyperparameter Training: EnSi FastText}
    \label{tab:alpha_ensi_fasttext}
    \begin{tabular}{|c|c|c|c|c|c|c|c|}
        \hline
        \textbf{SRC} & \textbf{-0.5} & \textbf{-0.25} & \textbf{-0.15} & \textbf{0} & \textbf{0.15} & \textbf{0.25} & \textbf{0.5} \\ \hline
        \textbf{TRG} &  &  &  &  &  &  & \\ \hline
        \textbf{-0.5} & 0.09 & 0 & 0 & 0.09 & 0 & 0.18 & 0.09 \\ \hline
        \textbf{-0.25} & 0 & 0.45 & 0 & 0.27 & 20.07 & 20.79 & 0.27 \\ \hline
        \textbf{-0.15} & 0 & 0.36 & 22.39 & 23.82 & 24.8 & 22.84 & 0.09 \\ \hline
        \textbf{0} & 0 & 4.46 & 23.64 & 25.69 & 27.48 & 26.05 & 6.07 \\ \hline
        \textbf{0.15} & 0.09 & 0.62 & 25.87 & 27.74 & 28.37 & 27.48 & 9.81 \\ \hline
        \textbf{0.25} & 0 & 23.91 & 25.16 & \textbf{28.81} & 27.48 & 27.65 & 22.75 \\ \hline
        \textbf{0.5} & 0.18 & 20.79 & 22.66 & 24.62 & 25.51 & 25.78 & 22.48 \\ \hline
    \end{tabular}
    
\end{table}

\begin{table}[h]
    \centering
    \caption{Alpha Hyperparameter Training: EnSi Word2Vec}
    \label{tab:alpha_ensi_w2v}
    \begin{tabular}{|c|c|c|c|c|c|c|c|}
        \hline
        \textbf{SRC} & \textbf{-0.5} & \textbf{-0.25} & \textbf{-0.15} & \textbf{0} &  \textbf{0.15} & \textbf{0.25} & \textbf{0.5} \\ \hline
        \textbf{TRG} &  &  &  &  &  &  & \\ \hline
        \textbf{-0.5} & 0.18 & 0.09 & 0 & 0 & 0.09 & 0 & 0 \\ \hline
        \textbf{-0.25} & 0 & 1.52 & 26.67 & 28.55 & 12.58 & 0.18 & 16.77 \\ \hline
        \textbf{-0.15} & 0.09 & 0.54 & 28.72 & 30.15 & 4.82 & 0.18 & 19.18 \\ \hline
        \textbf{0} & 0.45 & 29.44 & 32.11 & 32.56 & 31.67 & 30.51 & 0.54 \\ \hline
        \textbf{0.15} & 0.27 & 30.78 & 21.14 & 33.1 & 32.92 & 21.5 & 0.09 \\ \hline
        \textbf{0.25} & 0.09 & 29.62 & 32.02 & 32.83 & 33.18 & 31.58 & 26.76 \\ \hline
        \textbf{0.5} & 0 & 24 & 27.3 & 29.08 & 29.62 & 29.53 & 27.12 \\ \hline
    \end{tabular}
    
\end{table}

\begin{table}[h]
    \centering
    \caption{Alpha Hyperparameter Training: EnTa FastText}
    \label{tab:alpha_enta_fasttext}
    \begin{tabular}{|c|c|c|c|c|c|c|c|}
        \hline
        \textbf{SRC} & \textbf{-0.5} & \textbf{-0.25} & \textbf{-0.15} & \textbf{0} & \textbf{0.15} & \textbf{0.25} & \textbf{0.5} \\ \hline
        \textbf{TRG} &  &  &  &  &  &  & \\ \hline
        \textbf{-0.5} & 0 & 0.22 & 0 & 0 & 0 & 0.11 & 0 \\ \hline
        \textbf{-0.25} & 0 & 0.11 & 0 & 0 & 0 & 0 & 0.11 \\ \hline
        \textbf{-0.15} & 0.11 & 0.11 & 0 & 0 & 0.11 & 0.34 & 0.11 \\ \hline
        \textbf{0} & 0 & 0 & 0 & 13.15 & 0.34 & 0.11 & 0.11 \\ \hline
        \textbf{0.15} & 0 & 0.11 & 0.11 & 0.12 & \textbf{14.04} & 0.11 & 0.11 \\ \hline
        \textbf{0.25} & 0.11 & 0.11 & 0.9 & 0.56 & 13.03 & 0 & 0.11 \\ \hline
        \textbf{0.5} & 0 & 0 & 4.16 & 0.9 & 8.09 & 0 & 0 \\ \hline
    \end{tabular}
    
\end{table}

\begin{table}[h]
    \centering
    \caption{Alpha Hyperparameter Training: EnTa Word2Vec}
    \label{tab:alpha_enta_w2v}
    \begin{tabular}{|c|c|c|c|c|c|c|c|}
        \hline
        \textbf{SRC} & \textbf{-0.5} & \textbf{-0.25} & \textbf{-0.15} & \textbf{0} & \textbf{0.15} & \textbf{0.25} & \textbf{0.5} \\ \hline
        \textbf{TRG} &  &  &  &  &  &  & \\ \hline
        \textbf{-0.5} & 0.11 & 0.11 & 0 & 0 & 0 & 0 & 0 \\ \hline
        \textbf{-0.25} & 0 & 0.22 & 12.92 & 13.71 & 0 & 0 & 6.85 \\ \hline
        \textbf{-0.15} & 0 & 11.8 & 2.36 & 15.17 & 15.39 & 12.81 & 9.21 \\ \hline
        \textbf{0} & 0.22 & 13.6 & 0.67 & 16.18 & \textbf{16.4} & 1.91 & 5.39 \\ \hline
        \textbf{0.15} & 0.11 & 0.11 & 0 & 15.39 & 2.92 & 14.27 & 11.24 \\ \hline
        \textbf{0.25} & 0 & 0.9 & 12.7 & 0.22 & 15.17 & 14.72 & 11.35 \\ \hline
        \textbf{0.5} & 0 & 5.62 & 7.87 & 9.44 & 3.6 & 10.67 & 9.78 \\ \hline
    \end{tabular}
    
\end{table}

\begin{table}[h]
    \centering
    \caption{Alpha Hyperparameter Training: EnPa FastText}
    \label{tab:alpha_enpj_fasttext}
    \begin{tabular}{|c|c|c|c|c|c|c|c|}
        \hline
        \textbf{SRC} & \textbf{-0.5} & \textbf{-0.25} & \textbf{-0.15} & \textbf{0} & \textbf{0.15} & \textbf{0.25} & \textbf{0.5} \\ \hline
        \textbf{TRG} &  &  &  &  &  &  & \\ \hline
        \textbf{-0.5} & 0 & 0.08 & 0 & 0 & 0 & 0 & 0 \\ \hline
        \textbf{-0.25} & 0 & 14.66 & 15.05 & 15.44 & 15.52 & 14.58 & 0 \\ \hline
        \textbf{-0.15} & 0.16 & 14.81 & 15.36 & 15.75 & 15.44 & 14.73 & 0.16 \\ \hline
        \textbf{0} & 0 & 15.13 & 15.28 & 15.6 & 15.67 & 15.05 & 12.23 \\ \hline
        \textbf{0.15} & 0 & 15.05 & 15.99 & 15.67 & 15.44 & 15.44 & 12.38 \\ \hline
        \textbf{0.25} & 0 & 15.75 & \textbf{16.22} & 15.83 & 14.81 & 14.26 & 12.46 \\ \hline
        \textbf{0.5} & 0 & 0.08 & 0.24 & 14.34 & 14.03 & 13.64 & 11.44 \\ \hline
    \end{tabular}
    
\end{table}

\begin{table}[h]
    \centering
    \caption{Alpha Hyperparameter Training: EnPa Word2Vec}
    \label{tab:alpha_enpj_w2v}
    \begin{tabular}{|c|c|c|c|c|c|c|c|}
        \hline
        \textbf{SRC} & \textbf{-0.5} & \textbf{-0.25} & \textbf{-0.15} & \textbf{0} & \textbf{0.15} & \textbf{0.25} & \textbf{0.5} \\ \hline
        \textbf{TRG} &  &  &  &  &  &  & \\ \hline
        \textbf{-0.5} & 0 & 0 & 0.16 & 0.08 & 0 & 0 & 0 \\ \hline
        \textbf{-0.25} & 0 & 14.42 & 14.34 & 14.34 & 14.18 & 13.95 & 0 \\ \hline
        \textbf{-0.15} & 0 & 14.97 & 14.97 & 15.2 & 14.58 & 14.34 & 0 \\ \hline
        \textbf{0} & 0 & \textbf{15.36} & 14.97 & 14.81 & 14.81 & 14.81 & 12.54 \\ \hline
        \textbf{0.15} & 0 & 14.97 & 15.28 & 15.2 & 14.66 & 14.5 & 12.54 \\ \hline
        \textbf{0.25} & 0 & 0.08 & 0.31 & 15.2 & 14.26 & 14.18 & 12.46 \\ \hline
        \textbf{0.5} & 0 & 0 & 0 & 0 & 13.48 & 13.01 & 11.36 \\ \hline
    \end{tabular}
    
\end{table}

\section{\textbf{Frequency Thresholds}}
\label{sec:appendix_threshold}
Tables~\ref{tab:min_freq_enta} to~\ref{tab:min_freq_ensi} illustrate the impact of minimum frequency thresholds on accuracy for the EnTa, EnPa, and EnSi datasets.

\begin{table}[h]
    \centering
    \caption{UVecMap Results for Different Frequency Thresholds and their Impact on Accuracy(pr@1) and Coverage (EnTa Dataset)}
    \label{tab:min_freq_enta}
    \setlength{\tabcolsep}{2pt} 
    \begin{tabular}{|c|c|c|c|}
        \hline
        \textbf{Min\_Freq} & \multicolumn{2}{|c|}{\textbf{Accuracy}} & \textbf{Coverage} \\ \hline
        & \textbf{W2V} & \textbf{FASTTEXT} &  \\ \hline
        \textbf{2} & 1.14 & 1.23 & 98.87\% \\ \hline
        \textbf{4} & 12.78 & 0.1 & 86.14\% \\ \hline
        \textbf{6} & \textbf{15.39} & \textbf{12.81} & \textbf{77.12\%} \\ \hline
        \textbf{8} & 16.54 & 15.2 & 70.71\% \\ \hline
        \textbf{10} & 20.16 & 17.28 & 66.2\% \\ \hline
    \end{tabular}       
\end{table}

\begin{table}[h]
    \centering
    \caption{UVecMap Results for Different Frequency Thresholds and their Impact on Accuracy(pr@1) and Coverage (EnPa Dataset)}
    \label{tab:min_freq_enpj}
    \setlength{\tabcolsep}{2pt} 
    \begin{tabular}{|c|c|c|c|}
        \hline
        \textbf{Min\_Freq} & \multicolumn{2}{|c|}{\textbf{Accuracy}} & \textbf{Coverage} \\ \hline
        & \textbf{W2V} & \textbf{FASTTEXT} &  \\ \hline
        \textbf{2} & 12.53 & 15.22 & 98.78\% \\ \hline
        \textbf{4} & 13.94 & 14.62 & 95.47\% \\ \hline
        \textbf{6} & \textbf{15.13} & \textbf{15.05} & \textbf{91.8\%} \\ \hline
        \textbf{8} & 15.87 & 15.13 & 88.42\% \\ \hline
        \textbf{10} & 16.46 & 16.37 & 85.25\% \\ \hline
    \end{tabular}       
\end{table}

\begin{table}[h]
    \centering
    \caption{UVecMap Results for Different Frequency Thresholds and their Impact on Accuracy(pr@1) and Coverage (EnSi Dataset)}
    \label{tab:min_freq_ensi}
    \setlength{\tabcolsep}{2pt} 
    \begin{tabular}{|c|c|c|c|}
        \hline
        \textbf{Min\_Freq} & \multicolumn{2}{|c|}{\textbf{Accuracy}} & \textbf{Coverage} \\ \hline
        & \textbf{W2V} & \textbf{FASTTEXT} &  \\ \hline
        \textbf{2} & 19.42 & 8.82 & 93.02\% \\ \hline
        \textbf{4} & 20.03 & 20.63 & 85.07\% \\ \hline
        \textbf{6} & 24.25 & 25 & 77.39\% \\ \hline
        \textbf{8} & \textbf{27.48} & \textbf{31.49} & \textbf{71.81\%} \\ \hline
        \textbf{10} & 30.91 & 32.53 & 63.42\% \\ \hline
    \end{tabular}       
\end{table}

\end{document}